%% file: main.tex
\documentclass[a4paper,11pt]{article}
\usepackage{jinstpub} 
\usepackage{multirow}
\usepackage{lineno}
\usepackage{hyperref}
\usepackage{tikz}
\usetikzlibrary{shapes.geometric, arrows}
\usepackage{listings}
\usepackage{glossaries}
\usepackage{graphicx}
\usepackage{subcaption}
\usepackage{comment}
\usepackage{enumitem}

\newcommand{\addComment}[2]{
  \expandafter\newcommand\csname #1\endcsname[1]{{\bf \color{#2} \capitalisewords{#1}:\,##1}}
  \expandafter\newcommand\csname #1cor\endcsname[2]{{\color{#2} \capitalisewords{#1}:\,\st{##1}{\bf ##2}}}
  \expandafter\newcommand\csname #1color\endcsname{#2}
}

\addComment{cris}{blue} 
\addComment{karthik}{red}

\input{glossary}

\title{Towards a RAG-based Summarization for the Electron Ion
Collider}

\author[]{Karthik Suresh*,}
\author[]{Neeltje Kackar,}
\author[]{Luke Schleck,}
\author[]{Cristiano Fanelli}

\affiliation[]{College of William \& Mary \\ Williamsburg, VA, USA}

\emailAdd{ksuresh@wm.edu}

\abstract{
The complexity and sheer volume of information—encompassing documents, papers, data, and other resources—from large-scale experiments demand significant time and effort to navigate, making the task of accessing and utilizing these varied forms of information daunting, particularly for new collaborators and early-career scientists.
To tackle this issue, a Retrieval Augmented Generation (RAG)-based Summarization AI for EIC (RAGS4EIC) is under development. This AI-Agent not only condenses information but also effectively references relevant responses, offering substantial advantages for collaborators. Our project involves a two-step approach: first, querying a comprehensive vector database containing all pertinent experiment information; second, utilizing a Large Language Model (LLM) to generate concise summaries enriched with citations based on user queries and retrieved data. We describe the evaluation methods that use RAG assessments (RAGAs) scoring mechanisms to assess the effectiveness of responses. Furthermore, we describe the concept of prompt template based instruction-tuning which provides flexibility and accuracy in summarization. Importantly, the implementation relies on LangChain~\cite{lang-chain}, which serves as the foundation of our entire workflow. This integration ensures efficiency and scalability, facilitating smooth deployment and accessibility for various user groups within the Electron Ion Collider (EIC) community. This innovative AI-driven framework not only simplifies the understanding of vast datasets but also encourages collaborative participation, thereby empowering researchers. 
As a demonstration, a web application has been developed to explain each stage of the RAG Agent development in detail. The application can be accessed at \href{https://rags4eic-ai4eic.streamlit.app}{https://rags4eic-ai4eic.streamlit.app}\footnote{A tagged version of the source code can be found in \href{https://github.com/ai4eic/EIC-RAG-Project/releases/tag/AI4EIC2023_PROCEEDING}{https://github.com/ai4eic/EIC-RAG-Project/releases/tag/AI4EIC2023\_PROCEEDING}}.
}


\begin{document}
\maketitle
\flushbottom

\section{Background}
\label{sec:bkg}

\paragraph{The Electron Ion Collider (EIC):}
The EIC is the next generation large-scale scientific project that is anticipated to become operational within the next decade \cite{EIC:Yellow-Report}. %
The EIC User group is a fast-evolving community consisting of more than 1,400 physicists from over 38 countries around the world. Book keeping has been a significant challenge in such large collaborative experiments; specifically, in the case of large-scale physics experiments, one faces significant challenges in coordinating information curation between numerous working groups within the collaboration. 

New collaborators in such a large-scale experiment often feel overwhelmed when they start. For example, when conducting experiments and collecting data, the institutions involved in the collaboration need to comprehend and review a vast amount of documentation during their respective shifts, which can be daunting for the beginner. In this context, having a virtual assistant available to support shift takers would be immensely beneficial.


\paragraph{Fine tuning of Large Language Models (LLMs):}

Fine-tuning an LLM involves refining its abilities and performance in specific tasks or domains by training it further in domain-specific datasets after pretraining, improving effectiveness without retraining the entire model \cite{sun2020finetune,Mosin_2023}. However, conventional fine-tuning requires substantial computational power and time, posing challenges, especially for extensive models such as GPT-3 (175-B parameters)\cite{brown2020language} and Meta LLaMA2 (13-B parameters) \cite{touvron2023llama}. Despite the advent of Low-Rank Adaptation (LoRA) which enables efficient fine-tuning, even on consumer-grade GPUs, fine-tuning remains computationally intensive. Integrating strategies like in-context learning and chain-of-thought techniques \cite{dong2023survey} with a live knowledge repository forms the concept of Retrieval Augmented Generation (RAG), reducing hallucinations and anchoring the LLMs to reality

\section{Retrieval Augmented Generation pipeline}

\begin{figure}[!h]
    \centering
    \includegraphics[scale = 0.225]{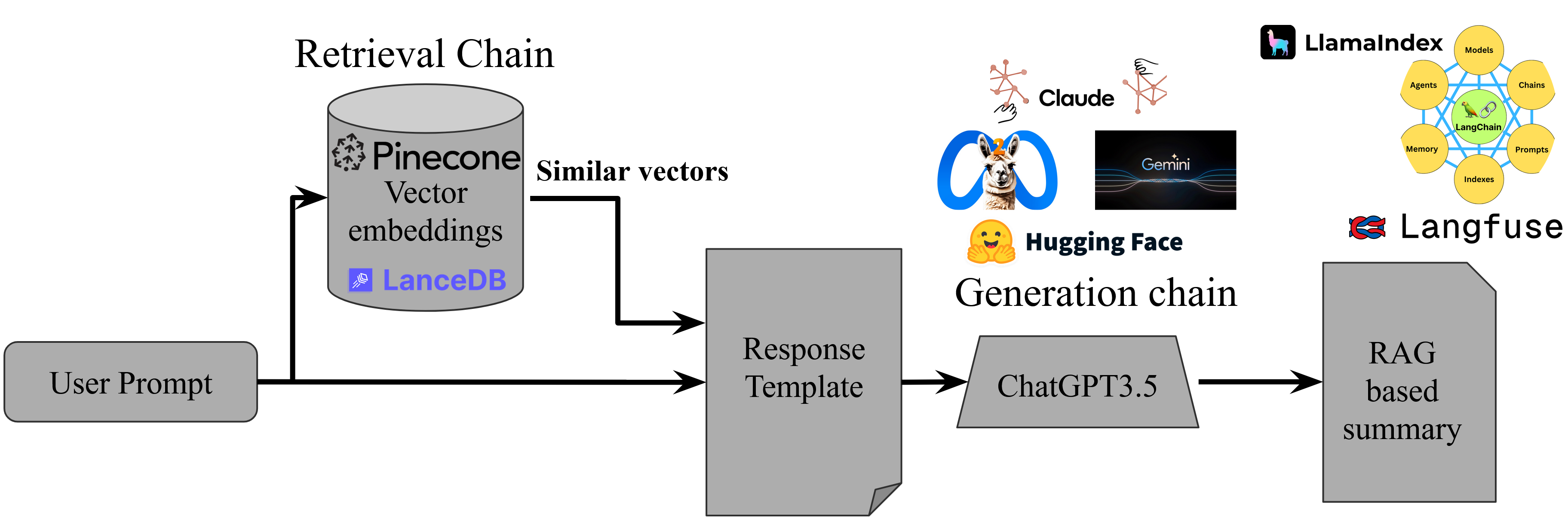}
    \caption{
    A Naive RAG Agent involves the user inputting a prompt to the agent. The agent leverages an external knowledge base to gather more details related to the query.  The information is enhanced using a predefined response template, which is fed in
as an input to a Frozen LLM such as OpenAI's GPT, Anthropic's Claude or Meta's LLaMA2 for summarization of the result. The RAG Agent operates within a LLM engineering platform such as LangChain, LangFuse or LlamaIndex frameworks.}
    \label{fig:RAG-inference}
\end{figure}

The concept of augmented retrieval generation was first proposed in late 2020 as a general purpose framework that combined parametric and nonparametric pre-trained memory for language generation \cite{lewis2021retrievalaugmented}. It showed that RAG models can achieve state-of-the-art results in various knowledge-intensive NLP tasks, such as open domain question answering, abstractive question answering, Jeopardy question generation, and fact verification. Since its proposal, RAG methods have continuously evolved  to achieve superior performance in grounding LLM to truth and reducing hallucinations. Different RAG methods that involve the use of LLMs have been proven to be more effective in recent work \cite{gao2024retrievalaugmented, asai2023selfrag}. The workflow of a Naive RAG Agent is represented in  Fig.~\ref{fig:RAG-inference}.

The objective of this study is to design and deploy a proof-of-concept RAG Agent specifically designed to address questions regarding the EIC science based on a small knowledge base. The purpose of this agent is to answer user questions and provide references for the information retrieved, guaranteeing its trustworthiness and authenticity. To build an RAG Agent, the first phase includes setting up a knowledge base and then extracting valuable insights from this repository of knowledge.

\section{Creation of Knowledge base}

To establish the knowledge base, the selection of the source knowledge is the initial step, which is then followed by the ingestion process. During this stage, data are converted into numerical vectors and saved to facilitate quick retrieval when conducting inference.

\paragraph{EIC arXiv dataset:}

We built a selected collection of papers pertaining to the Electron Ion Collider from the arXiv database. These articles specifically cover the period of time after 2021. This approach was taken to ensure that the language model (GPT 3.5) did not possess prior knowledge of the content contained within these articles, as its training data only extend up to September 2021 \cite{ray2023chatgpt}. Consequently, any new information produced after this date is unlikely to be implicitly stored in the model weights. The complete list of arXiv sources utilized for this study can be accessed at \href{https://github.com/ai4eic/EIC-RAG-Project/blob/main/ingestion/ARXIV_SOURCES.info}{ARXIV\_SOURCES.info}.

\paragraph{Ingestion:}

\begin{figure}[!h]
    \centering
    \includegraphics[scale = 0.125]{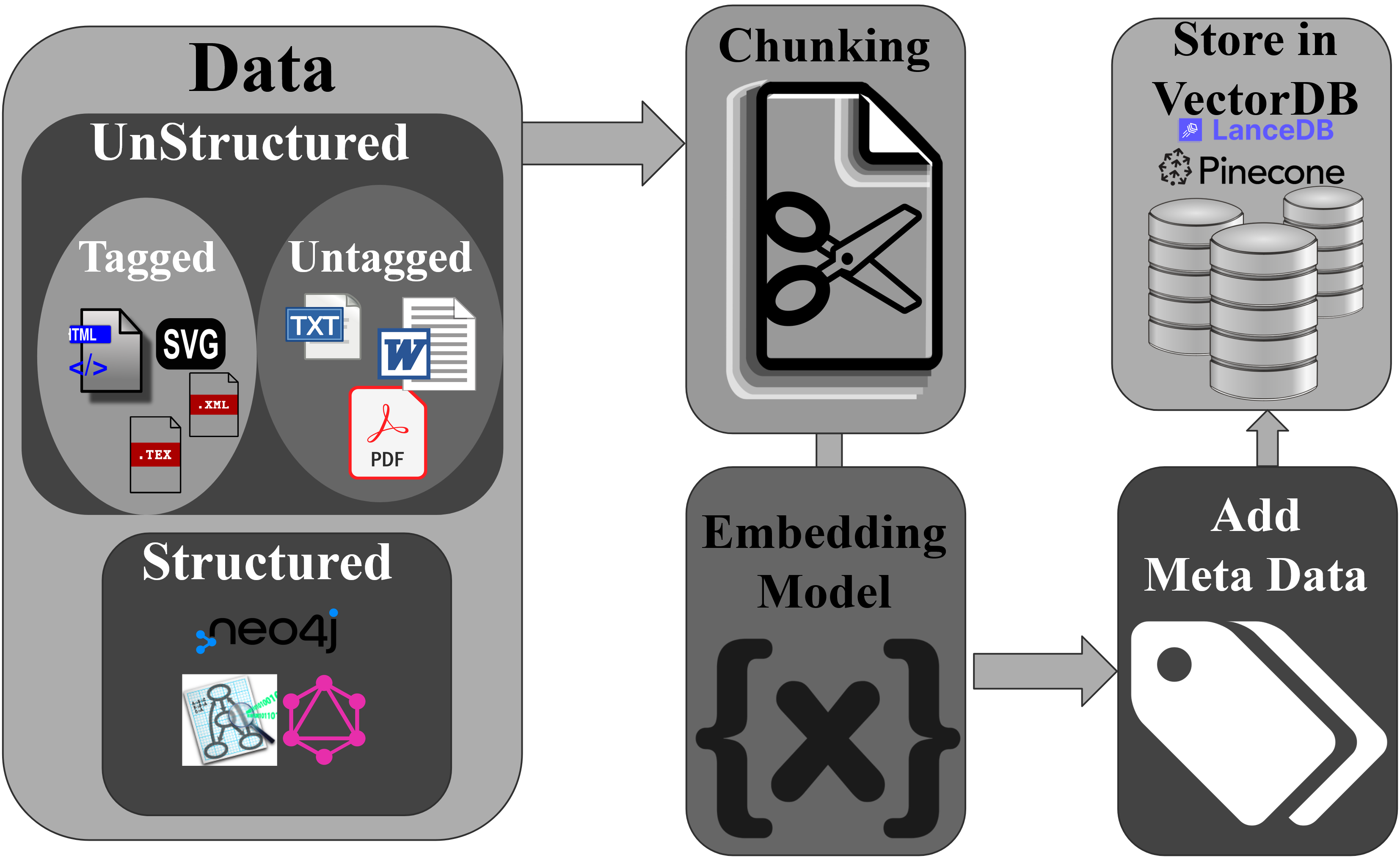}
    \caption{%
    Creating a knowledge base involves ingesting data in varied formats. EIC data, often unstructured yet tagged (e.g., wikis, run logs), includes untagged PDFs. Despite Optical Character Recognition (OCR) and Deep Learning models for text conversion, extracting figures and images from PDFs remains challenging, complicating the development of a multi-modal pipeline.
    }
    \label{fig:ingestion}
\end{figure}

Ingestion of data is a vital procedure for a Retrieval-Augmented Generation (RAG) pipeline. This procedure can be divided into three primary stages as described in Fig.~\ref{fig:ingestion}.

\begin{itemize}[leftmargin=10pt,noitemsep,topsep=0pt,parsep=0pt,partopsep=0pt]
    \item \textit{Chunking}: To begin the ingestion process, the initial step involves handling raw data. These raw data can be in various forms and have to be subsequently partitioned into manageable chunks or segments of semantic texts. The size of these pieces may depend on the specific given task. The fragmentation procedure helps to simplify the data, making it easier for the model to process the information efficiently. Nonetheless, it is crucial to know that chunking information may lead to incomplete data and loss of information.
    \item \textit{Encoding}: After fragmenting the information, various deep learning-based embedding methods such as \texttt{text-embedding-ada-002}, \texttt{BERT}, \texttt{seq2seq}, and \texttt{word2vec} \cite{mikolov2013efficient} are used to compress the information into high-dimensional vectors, thus quantifying them into numerical values that improve retrieval performance. These embedding models are usually shallow networks with a large number of nodes that transform semantic phrases into fixed-size numerical vectors. Therefore, two semantic phrases can be compared using a similarity metric and can be utilized in retrieval processes. Important measures to evaluate the similarity between distinct semantic segments include cosine similarity, Euclidean distance, and dot product. Cosine similarity is especially effective for semantic searches.
    \item \textit{Vector storage}: Vectorized segments are saved in a persistent storage system. Local database options provide low latency, but limit availability during application deployment. Therefore, the use of local VectorDB solutions is preferred during the prototyping stage. Once prototyping has been sufficiently refined, cloud-based storage provides greater accessibility and is suitable for small to medium-scale applications.
\end{itemize}

Table~\ref{tab:ingestion} displays a detailed summary of the ingestion process for the Advanced RAG Agent that was developed.
%
%
\begin{table}[h]
\centering
\resizebox{\textwidth}{!}{%
\begin{tabular}{|c|cc|c|l}
\cline{1-4}
\textbf{Information}               & \multicolumn{2}{c|}{\textbf{Details}}                                                                                                             & \textbf{Remarks}                                                                                                                                      &  \\ \cline{1-4}
\multirow{2}{*}{Dataset}           & \multicolumn{1}{c|}{\multirow{2}{*}{\begin{tabular}[c]{@{}c@{}}ARXIV papers\\ (dated after 2021)\end{tabular}}} & PDF Files                       & The results discussed correspond to the PDF Files                                                                                                     &  \\ \cline{3-4}
                                   & \multicolumn{1}{c|}{}                                                                                           & Source \texttt{\texttt{.tex}} files               & \begin{tabular}[c]{@{}c@{}}Custom method of chunking will allow \\ to include key figures during response\end{tabular}                                &  \\ \cline{1-4}
\multirow{5}{*}{Chunking}          & \multicolumn{1}{c|}{PDF Reader}                                                                                 & PyPDF2                          & Figures are discarded inherently in this reader                                                                                                       &  \\ \cline{2-4}
                                   & \multicolumn{1}{c|}{LangChain text splitter}                                                                    & RecursiveCharacterTextSplitter  & More details in \cite{lang-chain}                                                                                                    &  \\ \cline{2-4}
                                   & \multicolumn{1}{c|}{Latex Reader}                                                                               & LatexSplitter                   & \begin{tabular}[c]{@{}c@{}}Reads \texttt{.tex} files based on tags. Work continues to better chunk the information\end{tabular}                                &  \\ \cline{2-4}
                                   & \multicolumn{1}{c|}{Text chunk size}                                                                            & 120 Characters                  & \begin{tabular}[c]{@{}c@{}}Additional variation \\ 60 chars and 180 chars are being explored\end{tabular}                                                       &  \\ \cline{2-4}
                                   & \multicolumn{1}{c|}{overlap size}                                                                               & 10 Characters                   & \begin{tabular}[c]{@{}c@{}}With chunk size fixed at 120 \\ a variation of 20 characters was explored\end{tabular}                                     &  \\ \cline{1-4}
\multirow{3}{*}{Vector Encodings}  & \multicolumn{1}{c|}{Embedding model}                                                                            & OpenAI's text-embedding-ada-002 & \begin{tabular}[c]{@{}c@{}}No other embedding models were explored in this study\end{tabular}                                                      &  \\ \cline{2-4}
                                   & \multicolumn{1}{c|}{Output size vector}                                                                         & 1536                            &                                                                                                                                                       &  \\ \cline{2-4}
                                   & \multicolumn{1}{c|}{metadata}                                                                                   & arxiv-id, primary category & This is used during retrieval for citation purpose                                                                                                    &  \\ \cline{1-4}
\multirow{2}{*}{Vector DB Storage} & \multicolumn{1}{c|}{Vector DB Service}                                                                          & PineCone                        & \begin{tabular}[c]{@{}c@{}}Other local databases such as LanceDB and ChromaDB were explored\end{tabular}                                             &  \\ \cline{2-4}
                                   & \multicolumn{1}{c|}{Similarity metric for retrieval}                                                            & Cosine Similarity               & \begin{tabular}[c]{@{}c@{}}Maximal Marginal Relevance (MMR) metric was explored. \\ Option to be used during inference.\end{tabular} &  \\ \cline{1-4}
\end{tabular}%
}
\caption{Table presenting the specifics of the ingestion pipeline for the created RAG Agent. Various methodological investigations have been carried out and are currently in progress, influencing the RAG Agent's performance. PineCone VectorDatabase was selected for its simplicity in deployment and utilization throughout the ingestion and inference stages. More details on the ingestion process can be found in the tagged version of the source code on GitHub.
}
\label{tab:ingestion}
\end{table}
It is crucial to emphasize that each entry in Table.~\ref{tab:ingestion} has an impact on the precision of vector retrieval. The cosine similarity metric is utilized to assess the degree of similarity between two sets of semantic information. The efficacy of this metric is closely related to the size of the data segments, with a reduction in the effectiveness of cosine similarity as the segment size increases. On the contrary, if the segment sizes are too small, there is the possibility of losing the semantic relevance of the data segments. 
Optimizing parameters is crucial for enhancing vector retrieval efficiency, with performance measured by retrieval latency and similarity metrics, involving inherent trade-offs and objectives. As this pilot study aims to demonstrate the feasibility of RAG for the EIC, detailed optimization is deferred to future work.

\section{Inference}

\begin{figure}[!h]
    \centering
    \includegraphics[scale = 0.25]{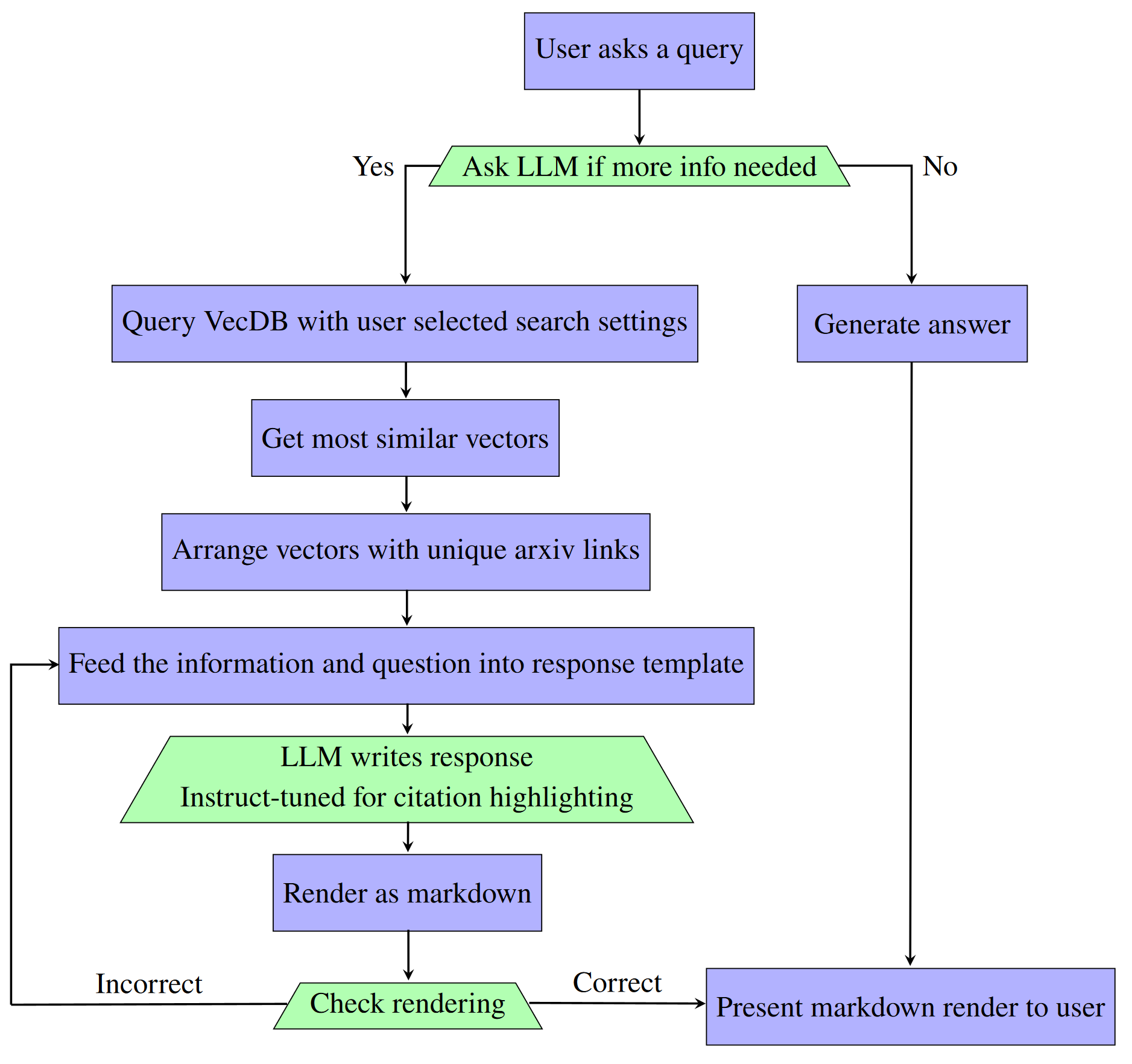}
    \caption{
    When a user submits a question and selects both the metric and search configuration for vector searching, the query first passes through a decision chain. The LLM evaluates whether to consult the knowledge base for the answer. If more information is needed, it searches the vector database for additional context and sources. 
    This information is processed through a fine-tuned template to craft a response with citations.
    The LLM then ensures syntax accuracy, delivering a GitHub markdown formatted answer. ChatGPT was used for this workflow with GPT-3.5 (\texttt{gpt-3.5-turbo-1106}) as the LLM running on the backend.
    } 
    \label{fig:rag-inference-pipeline}
\end{figure}

Once the knowledge base is ingested and stored in a vector database, the RAG inference pipeline can be built. In the current literature, there are broadly three types of RAG-based pipelines, and they are summarized in detail in \cite{gao2024retrievalaugmented}. For this study, a straightforward Advanced RAG pipeline is followed. The routing logic from the user question to the final response is shown in Fig.~\ref{fig:rag-inference-pipeline}.
Such an RAG Agent naturally requires to have a chat-based interface to use it efficiently; therefore, considerable efforts are being made to build a web-based chat application with detailed tracing using LangSmith \cite{lang-chain}. 
Key components of the information retrieval phase of the RAG Agent include:
\begin{itemize}[leftmargin=10pt,noitemsep,topsep=0pt,parsep=0pt,partopsep=0pt]
    \item \textbf{Similarity measures:} Users can choose between two sets of similarity metrics when utilizing data from VectorDB: (i) Cosine similarity and (ii) Maximal Marginal Relevance. The latter is especially beneficial for Question and Answer tasks, as it aids in reducing redundant entries. In addition, users can select the total maximum number of vectors to retrieve from the VectorDB.
    \item \textbf{Instruction tuned prompts:} To enhance the alignment of LLM for particular tasks, contextual learning techniques are used. Prompts that contain sample responses have been shown to improve summary performance. The prompt templates utilized for this study are outlined in the tagged source code on the GitHub repository.
\end{itemize}

\noindent An example of the application's inference is presented in Appendix \ref{sec:appendix}. 

\section{Evaluation of the EIC-RAG}

To assess the effectiveness of the RAG Agent, it is necessary to create a dataset comprised of series of questions along with their optimal answers and sources. This process typically requires a significant amount of human resources due to the need for expertise in the field. The dataset selected for this research encompasses a variety of disciplines, ranging from \texttt{hep.ph} to \texttt{nucl-ex} to \texttt{ph-acc}\footnote{These are the arXiv categorical codes in physics}. This inherently requires domain expertise to create reliable QA datasets. However, to alleviate the requirement for domain-specific knowledge, an AI-driven approach was adopted to produce a high quality dataset. The created dataset consists of targeted questions with a complete response along with the source information.

\paragraph{LLM-assisted creation of benchmark datasets:}

The primary goal of developing this data set is to utilize an advanced language model such as GPT-4.0 to generate a synthetic data set for evaluation purposes. Each question in the dataset is associated with a clearly defined set of "claims". Each question is uniquely characterized by the number of claims it addresses, and the corresponding answers to these claims are produced, culminating in a comprehensive response. In order to better facilitate this process, a routine within the web application was created where ``annotators" can generate new Question and Answer pairs and add to datasets. The dataset creation process involves the following steps:
\begin{enumerate}[leftmargin=10pt,noitemsep,topsep=0pt,parsep=0pt,partopsep=0pt]
    \item 
    The ``annotator'' chooses an arXiv paper (with an option for a random, unexplored selection), the total questions to generate, and the claims per question. GPT-4.0 then processes the paper's contents using a template to produce formatted Question and Answer pairs.
    \item Each QA pair has a question with `N' claims and a detailed \texttt{json} object which has detailed information about the answers.
    The \texttt{json} object contains the number of claims in the questions, the individual claims, ideal response to each of the individual claims, and a complete response involving all the claims. Hence, this method of data generation provides a rich and high quality dataset for evaluation. This approach also guarantees that even someone new to the field can generate a dataset with the help of LLMs.
    \item Subsequently, the ``annotator'' reviews the solution, makes any required additions or adjustments to the generated datasets, and incorporates it into the current dataset collection. An illustration of sample generation is provided in Appendix \ref{sec:appendix}. 
\end{enumerate}

\noindent A benchmark dataset named AI4EIC2023\_DATASETS\footnote{A dataset consisting of 50 targeted questions with up to 3 claims per question can be found \href{https://smith.langchain.com/public/a0fb3ae5-c878-4626-b8e4-1c45cc5fa566/d?tab=1&paginationState=\%7B\%22pageIndex\%22\%3A0\%2C\%22pageSize\%22\%3A10\%7D}{here}.}  was generated and used to evaluate the performance of the RAG Agent. In all evaluations, the cosine similarity metric was used to identify the 20 most closely related vectors for retrieval. Two main types of evaluation were performed on the dataset, which are summarized below.

\paragraph{Performance of RAG agent on standard metrics:}

Each of the questions in the data set has a defined number of claims ($M$). Each of the questions has the source of the article, individual responses to the claims, and a complete ideal response. The Retrieval Augmented Generation Assessment, along with a brief description of the metric, is shown in Table~\ref{tab:general-performance}.\footnote{The RAG Agent output response is captured as a PDF document and can be found \href{https://github.com/ai4eic/EIC-RAG-Project/blob/main/streamlit_app/Resources/assets/AI4EIC-RAG\%20ChatBot\%20response\%20AI4EIC2023.pdf}{here}.} 

\begin{table}[h]
\centering
\resizebox{0.7\columnwidth}{!}{%
\begin{tabular}{|c|c|c|l}
\cline{1-3}
\textbf{Metric Name}    & \textbf{Definition}                                                                                              & \textbf{Score}   &  \\ \cline{1-3}
Output Renderer Frequency         & \begin{tabular}[c]{@{}c@{}}Frequency of correctly rendered \\ output response in the markdown format\end{tabular} & $78.0\% \pm 5.8\% $ &  \\ \cline{1-3}
Claim Recognition Rate  & $\text{CRR} = \frac{|\text{Number of answered claims}|}{|\text{Total number of claims in question}|}$ &   $96.4\% \pm 3.4\%$  &  \\ \cline{1-3}
Claim Accuracy Rate     & $\text{CAR} = \frac{|\text{Number of correctly answered claims for a question}|}{\text{Total number of recognized claims in question}}$&   $88.9\% \pm 8.3\%$  &  \\ \cline{1-3}
Source Citation Frequency      &  $\text{SCF} = \frac{|\text{Number of time question's source cited}|}{|\text{Total queries}|}$ &  $85.3\% \pm 5.0\%$  &  \\ \cline{1-3}
Hallucination Frequency & $\text{HF} = \frac{|\text{Number of hallucinations}|}{\text{Total queries}}$ &   $2\% \pm 2\%$  &  \\ \cline{1-3}
\end{tabular}%
}
\caption{Standard assessment of the RAG Agent. Frequency metrics are subject to errors based on binomial distributions, while Rate metrics' errors are determined by the standard deviations of their score distributions.}
\label{tab:general-performance}
\end{table}

\paragraph{LLM-based metrics for evaluation - RAGAs:} 
The RAGAs framework was created to evaluate the effectiveness of QA chatbots \cite{es2023ragas}. 
This evaluation framework offers a comprehensive assessment of the performance of the RAG pipeline.
A significant feature of the RAGA framework is its capability to adapt to the use of LLMs for evaluation without the need for reference annotations by human experts. This strategy not only simplifies the evaluation process but also improves scalability by reducing the dependence on manual annotations. However, RAGAs also have potential drawbacks, particularly in terms of their subjective nature in the evaluation. 
This subjectivity could lead to biases and inconsistencies in the evaluation process if not closely monitored. Therefore, a controlled application of RAGAs is crucial during the evaluation phase. The performance of the RAG Agent on the RAGAs score is summarized in Table~\ref{tab:ragas-performance}\footnote{LLM used by RAGAs to score is GPT4.0.}.
\begin{table}[h]
\centering
\resizebox{0.7\textwidth}{!}{%
\begin{tabular}{|c|c|c|l}
\cline{1-3}
\textbf{Metric Name}  & \textbf{Definition}                                                                                                             & \textbf{Score}      &  \\ \cline{1-3}
Faithfulness          & \begin{tabular}[c]{@{}c@{}}Fraction of correctly rendered \\ output response in the markdown format\end{tabular}                & $87.4\% \pm 5.5\% $ &  \\ \cline{1-3}
Context Relevenacy    & \begin{tabular}[c]{@{}c@{}}Relevancy of the retrieved context \\ to the question and generated answer\end{tabular}              & $61.4\% \pm 4.3\%$  &  \\ \cline{1-3}
Context Entity Recall &  $\text{CER} = \frac{|\text{claims in context} \cap \text{claims in ground truth}|}{|\text{claims in ground truth}|}$  & $98.7\% \pm 1.2\%$  &  \\ \cline{1-3}
Answer Relevance      & \begin{tabular}[c]{@{}c@{}}Measures relevancy of the generated answer \\ to the retrieved context and the question\end{tabular} &   $77.2\% \pm 2.3\%$     &  \\ \cline{1-3}
Answer Correctness    & \begin{tabular}[c]{@{}c@{}}Measures correctness of the \\ generated answer to the idea response\end{tabular}                    & $72.3\% \pm 2.4\%$  &  \\ \cline{1-3}
\end{tabular}%
}
\caption{Performance of the RAG Agent on RAGAs scores. The errors shown are statistically extracted from the respective score distribution}
\label{tab:ragas-performance}.
\end{table}
The principal observations and corresponding conclusions regarding the performance of the RAG Agent are delineated below. Additionally, the subsequent steps to enhance the efficacy of EIC-RAG are enumerated.:
\begin{itemize}[leftmargin=10pt,noitemsep,topsep=0pt,parsep=0pt,partopsep=0pt]
    \item 
    The routing logic is designed to rewrite responses for GitHub Markdown rendering. To enhance this, the route could be optimized through instruction tuning or employing a model specifically fine-tuned for GitHub Markdown response rewriting.
    \item 
    The RAG Agent significantly reduces hallucinations by grounding responses to the knowledge base, achieving its primary goal. Including explicit claims in questions has enhanced its performance. Ongoing studies are further evaluating questions with implicitly mentioned claims.
    \item The RAG Agent's ability to provide accurate responses to inquiries decreases significantly when dealing with questions that involve physics equations (including special LaTeX characters). Enhancements can be made by implementing more effective chunking strategies and refining the LLM to enhance its comprehension of physics equations.
    \item Context Relevancy and Answer Relevance strongly depend on the total number of retrieved contexts. Since a fixed number of contexts ($k = 20$) were retrieved for each of the questions during the evaluations and the responses are relatively short, this leaves inherently redundant information, which is reflected in these metrics. 
    \item 
    The RAG Agent's Answer Correctness metric, which assesses semantic similarity between generated responses and the ideal answer, tends to score responses lower than the Claim Accuracy Score metric, which measures the precision of identified claims.
\end{itemize}

\section{Conclusions}
A RAG Agent has been successfully developed as a pilot study case for the upcoming EIC. The RAG Agent has demonstrated its capability to condense data sets and information and provide trustworthy references with very few instances of hallucinations, enhancing the understanding and effective information retrieval for the EIC community. A complete web application has been developed from dataset creation to inference. Detailed tracing is currently necessary for better tracking of these chains using LangSmith. The study has successfully demonstrated strong motivation to continue developing modular RAG methods for effective information retrieval for EIC. 
The use of LLM for creating artificial data has been studied; however, there is a requirement to create datasets by experts in the field to create top-notch reference data for evaluating the RAG pipelines. By improving an LLM with these datasets, it is possible to improve the techniques of LLM-supported synthetic data generation, providing a scalable method to generate extensive benchmark datasets. The subsequent phase in this area involves engaging diverse domain specialists to begin using the web tool and efficiently gather high-quality data from their use. The software infrastructure for these efforts is nearing its mature phase.
Evaluation pipelines such as RAGAs have been utilized to assess the performance of the RAG Agent without the need for references. The evaluators have not undergone parameter optimization to prevent any impact on the scores obtained. 
Despite the inherent challenge of reproducibility, efforts are being made to address this issue by effectively tracing the logic chains with tools such as LangSmith in order to reproduce the results on references. 
Multiple systematic investigations are needed to improve the comprehension of how the performance of the RAG system is influenced by factors such as ingestion, retrieval chain, LLM (\textit{e.g.}, GPT3.5, GPT4.0, Claude, LLaMA2), and potentially other hyperparameters.

In conclusion, the developed RAGS4EIC Agent marks a significant advancement in EIC technology, showcasing considerable potential to become the cornerstone companion for EIC applications in the foreseeable future.

\section*{Acknowledgements}

K.S. is supported by the Office of Nuclear Physics of the U.S. Department of Energy under Grant Contract No. DE-SC0024625

\bibliography{biblio}

\appendix

\section{Appendix} \label{sec:appendix}

\begin{figure}[h]
    \centering
    \includegraphics[scale = 0.225]{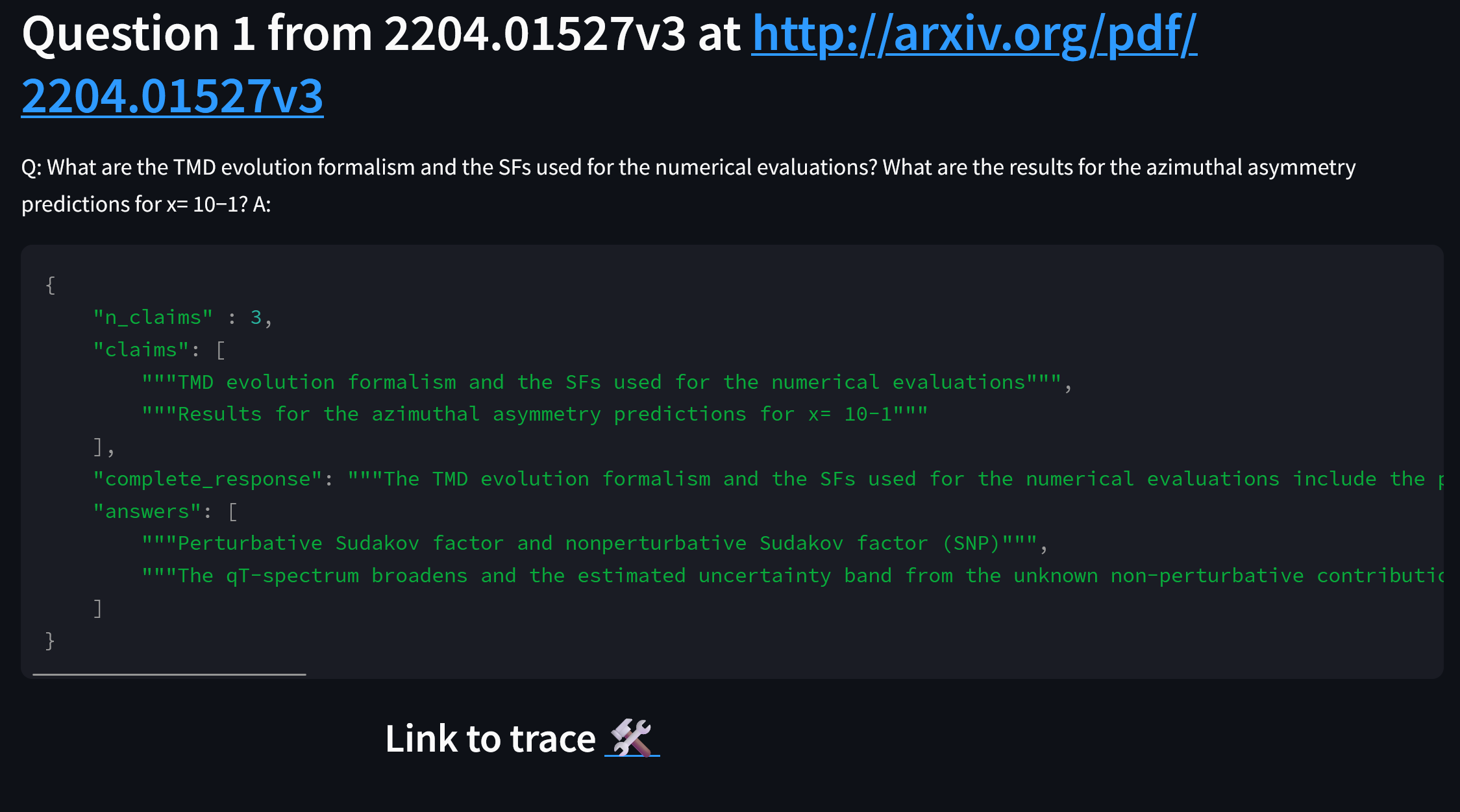}
    \caption{
    An example of creating an LLM-assisted Question \& Answer dataset. Within a web application, an ``annotator'' requests a question with three claims, generating a detailed JSON object that includes the number of claims, each claim, the ideal response for each claim, and a comprehensive response incorporating all claims. This process yields a rich, high-quality dataset for evaluation, enabling even newcomers to generate datasets with LLM assistance. The ``annotator'' can create multiple question-answer pairs and select specific ones for the dataset.
    }
    \label{fig:QA-DATAMODEL}
\end{figure}

\begin{figure}[h]
    \centering
    \begin{subfigure}[t]{\textwidth}
        \centering
        \includegraphics[scale=0.25]{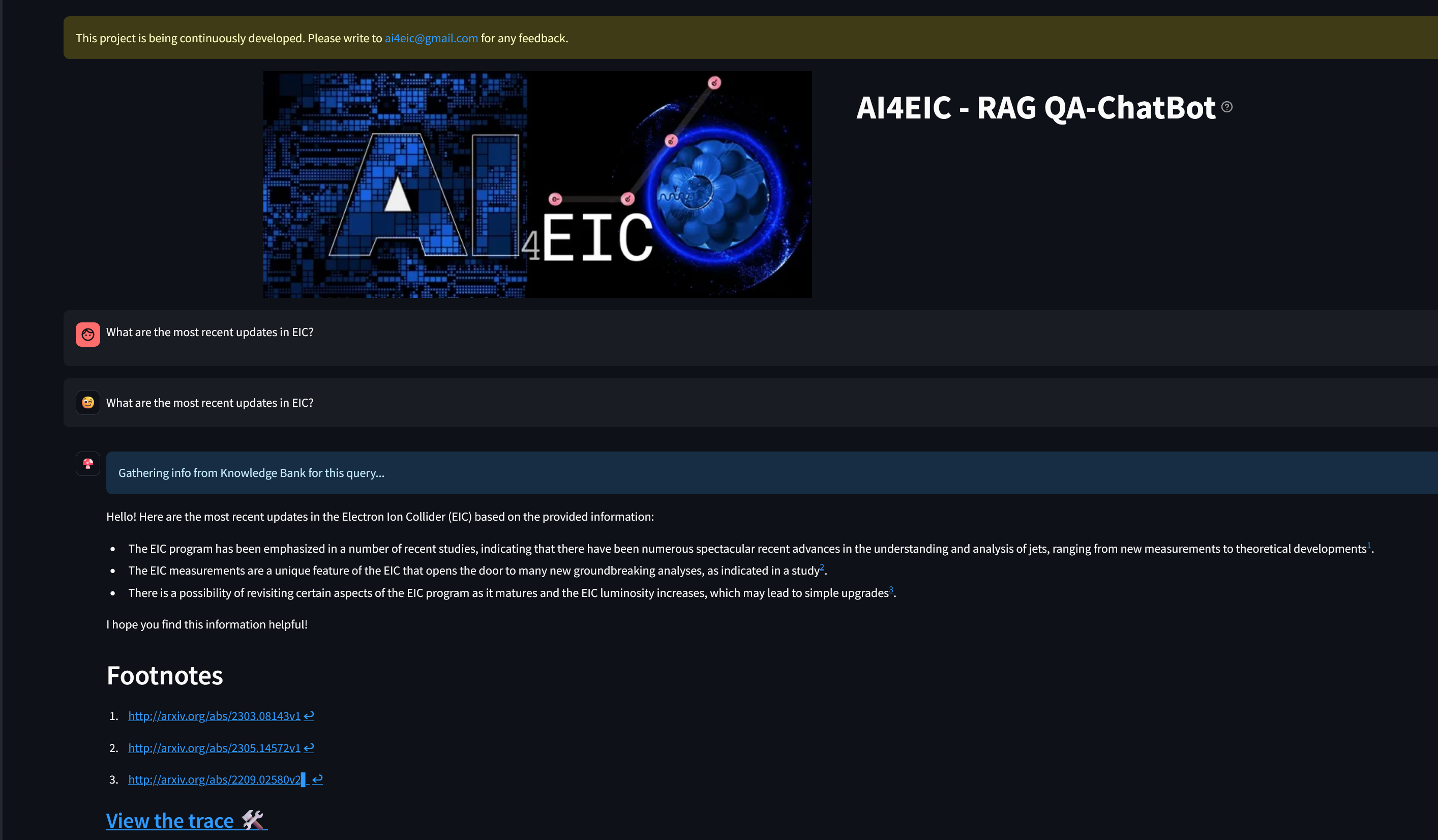}
        \caption{}
    \end{subfigure}
    \hfill
    \begin{subfigure}[b]{\textwidth}
        \centering
        \includegraphics[scale=0.25]{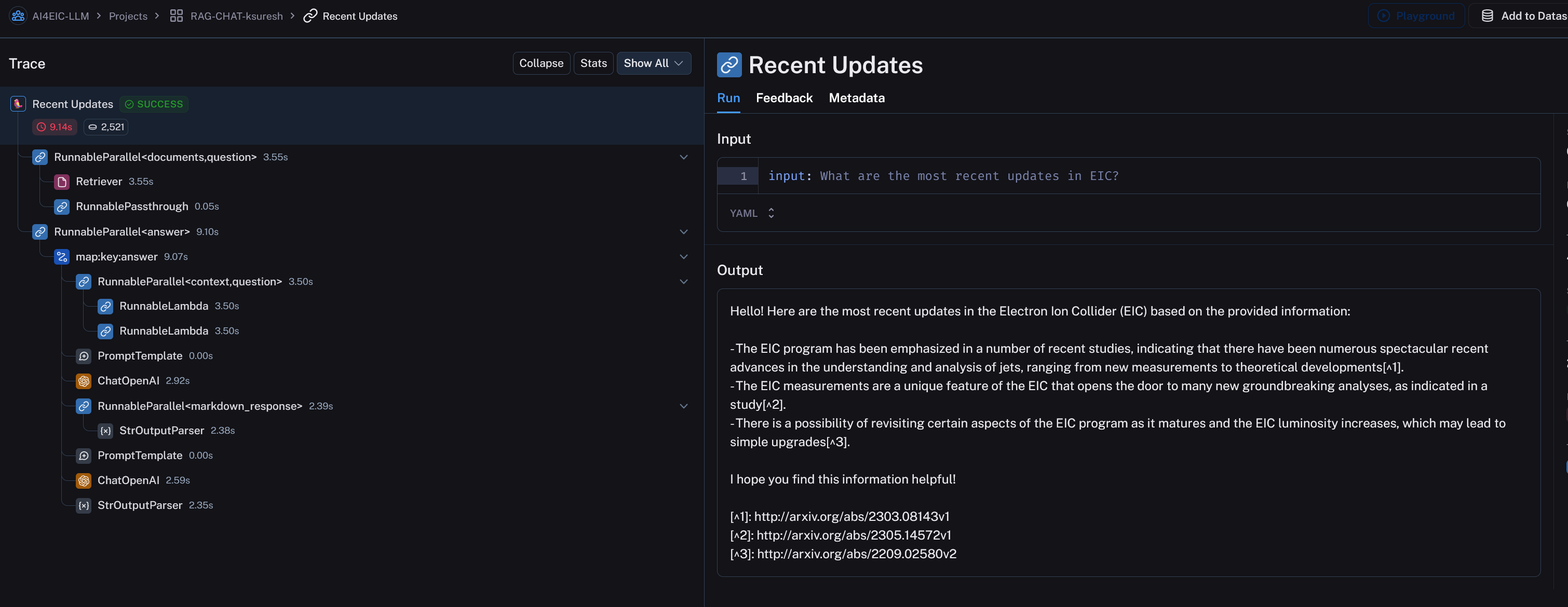}
        \caption{}
    \end{subfigure}
    \caption{
    The illustration outlines the RAGS4EIC inference process: (a) shows a user query and the Agent's decision on accessing the knowledge base, proceeding as detailed in Fig.~\ref{fig:rag-inference-pipeline}, and culminating in the output with a tracking link. (b) displays the query's chronological record in Langsmith, with logs maintained for up to 14 days before archival for possible future retrieval.
    }
    \label{fig:RAG-example-inference}
\end{figure}

This section shows in Fig. \ref{fig:QA-DATAMODEL} an example of how an LLM-assisted Question \& Answer dataset is created; and in Fig. \ref{fig:RAG-example-inference} the functioning of the RAGS4EIC inference process.

\end{document}

%% file: glossary.tex
\newacronym{rag}{RAG}{Retrieval Augmented Generation}

\newacronym{eic}{EIC}{Electron Ion Collider}